%% file: acl2021.tex
\documentclass[11pt,a4paper]{article}
\usepackage[hyperref]{acl2021}
\usepackage{times}
\usepackage{latexsym}

\usepackage{xcolor}

\usepackage{array}
\newcolumntype{H}{>{\setbox0=\hbox\bgroup}c<{\egroup}@{}}

\usepackage{microtype}

\aclfinalcopy 

\usepackage{latexsym}
\usepackage{cleveref}
\usepackage{multirow}
\usepackage{adjustbox}
\usepackage{booktabs}
\usepackage{graphicx}
\usepackage{url}
\usepackage{array}
\usepackage{subcaption}
\newcolumntype{H}{>{\setbox0=\hbox\bgroup}c<{\egroup}@{}}

\newcommand{\comment}[1]{}

\definecolor{ramiblue}{HTML}{007bd5}
\definecolor{ramigray}{HTML}{a9a9a9}
\definecolor{ramired}{HTML}{c11b24}

\definecolor{codegreen}{rgb}{0,0.6,0}
\definecolor{codegray}{rgb}{0.5,0.5,0.5}
\definecolor{codepurple}{rgb}{0.58,0,0.82}
\definecolor{paperred}{rgb}{.84,0,0,}
\definecolor{greybackground}{rgb}{.98,0.98,0.98}

\title{nmT5 - Is parallel data still relevant for pre-training massively multilingual language models?}

\author{Mihir Kale\thanks{\hspace{0.5em}Equal Contribution. Please direct correspondence to \{ \texttt{mihirkale}, \texttt{adisid}  \} \texttt{@google.com}}  \quad {\bf Aditya Siddhant}\footnotemark[1] \quad {\bf Noah Constant } \\ {\bf Melvin Johnson} \quad {\bf Rami Al-Rfou} \quad {\bf Linting Xue} \\
Google Research}
\date{}

\begin{document}
\maketitle
\begin{abstract}
Recently, mT5 - a massively multilingual version of T5 - leveraged a unified text-to-text format to attain state-of-the-art results on a wide variety of multilingual NLP tasks. In this paper, we investigate the impact of incorporating parallel data into mT5 pre-training. We find that multi-tasking language modeling with objectives such as machine translation during pre-training is a straightforward way to improve performance on downstream multilingual and cross-lingual tasks. However, the gains start to diminish as the model capacity increases, suggesting that parallel data might not be as essential for larger models. At the same time, even at larger model sizes, we find that pre-training with parallel data still provides benefits in the limited labelled data regime.
\end{abstract}

\section{Introduction}

Recent works have shown that cross-lingual transfer learning in pre-trained multilingual models such as mBERT \cite{devlin-etal-2019-bert} and XLM-R \cite{conneau2020unsupervised} could be improved further by using parallel data \cite{conneau2019advances,hu2020explicit,ouyang2020ernie,luo2020veco}. In this paper, we continue this line of work by improving the recent mT5 model \cite{xue2020mT5} by leveraging  parallel corpora. We experiment with several text-to-text objectives that incorporate parallel data (spanning 198 language pairs) into mT5 pre-training. Our key findings are summarized below:

\begin{itemize}
    \item In the regime of very small fine-tuning datasets, objectives with parallel data improve results significantly.
    \item The gain from using parallel data decreases as we scale up the size of the pre-trained model.
    \item Simple objectives based on neural machine translation (NMT) perform better than the traditionally employed ``translation language modeling'' (TLM) objective.
\end{itemize}


\begin{figure*}
    \centering
    \includegraphics[width=0.85\textwidth]{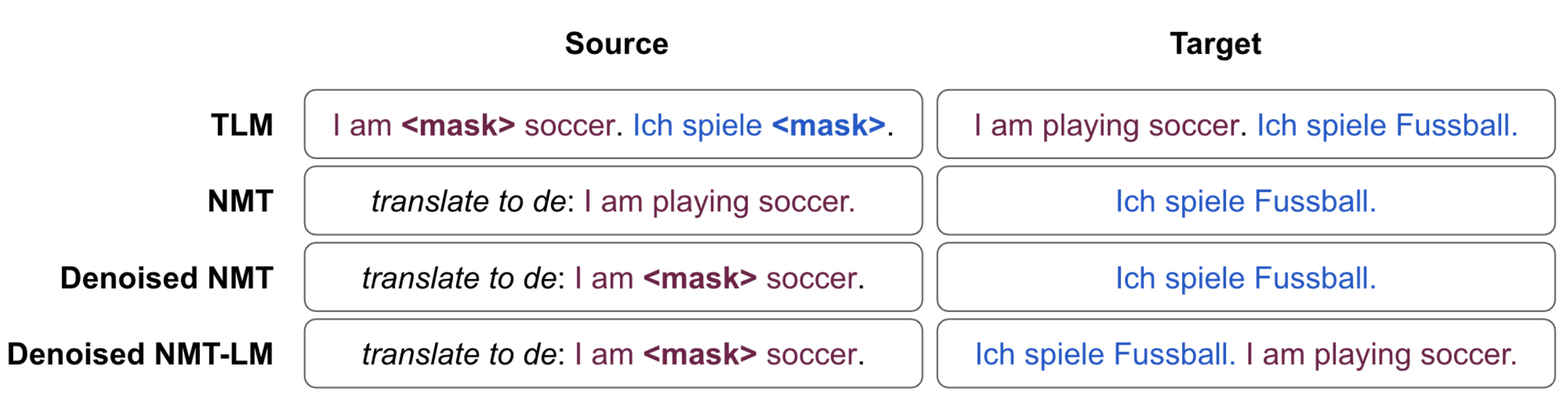}
    \caption{Example source and targets for different text-to-text style pre-training objectives incorporating parallel data. All objectives except TLM specify target language in the source sentence.}
    \label{fig:my_label}
\end{figure*}

\section{Method}
We focus on the mT5-Large model, which is a 24 layer encoder-decoder transformer model and has shown strong performance on a variety of cross-lingual benchmarks \cite{xue2020mT5}. Instead of training a new model from scratch, we start from the publicly available mT5-Large checkpoint - which has been trained for over 1 trillion tokens - and do a second stage pre-training with a mix of monolingual and parallel data.
\subsection{Objectives}
The mT5 - multilingual version of T5 \cite{2020t5} - series of models were pre-trained on a multilingual version of the C4 corpus with a masked language modeling ``span-corruption'' objective \cite{2020t5}, where the encoder is fed a chunk of text with random spans replaced with a mask token, and the decoder must reconstruct the masked-out tokens. One of their primary distinctions is the use of a unified ``text-to-text'' format for all text-based NLP problems.
\par In keeping with the text-to-text format, we experiment with the following objectives to incorporate parallel data into pre-training:

\begin{itemize}
    \item \textbf{TLM} - A text-to-text version of translation language modeling, proposed by \citet{conneau2019advances} and subsequently used in several prior works for encoder only pre-training. We trivially extend it to the encoder-decoder setting. 
    \item \textbf{NMT} - Standard machine translation. The input is the source text and the target is its translation. A language code is prefixed to the input to inform the model of the target language \cite{Johnson2017GooglesMN}.
    \item \textbf{Denoised-NMT} - Similar to NMT, but we additionally mask spans in the source sentence. The model must now learn to implicitly perform language modeling of the source language while translating into the target language.
    \item \textbf{Denoised-NMT+LM} - Similar to Denoised-NMT, but instead of implicit language modeling, the model must explicitly predict the source text in addition to the translation. The target is a concatenation of the translation and source sentence, while the input is the masked source sentence.
\end{itemize}
We refer to the model trained with the standard NMT objective as nmT5.

\section{Experiment Setup}

\label{sec:length}

\paragraph{Pre-training datasets}
For pre-training we use monolingual data from mC4 \cite{xue2020mT5} and parallel data from OPUS-100 \cite{zhang2020improving}. OPUS-100 is a dataset of 55M translations covering 100 languages (198 language pairs, either into or from English). The mC4 corpus consists of unlabeled web text covering 101 languages, of which 81 overlap with the OPUS-100 languages.
\paragraph{Fine-tuning datasets} For downstream evaluation, we use the following four tasks:

\begin{itemize}
    \item \textbf{TyDi QA} \cite{Clark2020TyDiQA} - The GoldP subtask, which corresponds to extractive question answering. The input is a passage and a question, with the answer being a span from the passage. 
    \item \textbf{MTOP} \cite{Li2020MTOPAC} - Multilingual Task-Oriented Parsing. The task is one of structured prediction,  where user queries must be parsed into a tree, capturing the domain, intent and slots.
     \item \textbf{WikiAnn NER} \cite{pan-etal-2019-cross} - Named entity recognition task covering 40 languages featured in the XTREME benchmark \cite{hu2020xtreme}. There are 4 categories of entities - location, person, organization and miscellaneous.
    \item \textbf{WikiLingua} \cite{Ladhak2020WikiLinguaAN} - A recently introduced \textit{cross-lingual} summarization dataset, where a document from an arbitrary language must be summarized in English. Since the dataset does not come with training and evaluation splits, we randomly create validation and test sets of 1000 examples each, and the rest of the data is used for training.
\end{itemize}
Table \ref{tab:dataset} lists further details of each dataset. Following \citet{xue2020mT5}, all tasks are cast into the text-to-text format. The evaluation for TyDi QA, MTOP and NER is done in the zero-shot setting, where the model is trained on the English data and evaluated on all languages. Since zero-shot cross-lingual language generation is much harder, for WikiLingua we train the model in a multilingual setting, using available training data for all languages.

\begin{table}[]
\resizebox{0.45\textwidth}{!}{\begin{tabular}{l|rrc}
\toprule
Dataset    & Langs       & Train size  & Setting\\ \midrule
TyDi QA    & 9            & 3.7K  & zero-shot      \\
MTOP       & 6           & 22K      & zero-shot    \\
WikiAnn NER & 40  & 20K & zero-shot\\
WikiLingua & 18        & 660K   & multilingual     \\ 
\bottomrule
\end{tabular}}
\caption{Statistics of datasets used in the paper.}\label{tab:dataset}
\end{table}

\begin{table*}[!htb]
\centering
\resizebox{0.75\textwidth}{!}{
\begin{tabular}{l|ccccc} 
\toprule
Model           & TyDi QA     & MTOP &  NER &  WikiLingua & Avg.  \\ 
 (Metric) & (F1/EM) & (EM) &  (F1) & (ROUGE-L) \\
\midrule
mT5             & 66.3 / 49.8 & 43.7 & 58.4 & 25.2 & 46.3       \\
+MLM (additional 100K steps)             & 71.3 / 55.6 & 48.6 & 59.9 & 26.1 & 49.5        \\
+MLM+TLM             & 71.1 / 54.6 & 48.6 & 61.4 & 26.1 & 49.7        \\
+MLM+NMT             & 75.1 / 60.1 & 57.7 & 61.4 & 27.4 & 53.5         \\
+MLM+denoised NMT    & 75.3 / 60.2 & 56.5 & 61.5 & 27.4 & 53.3        \\
+MLM+denoised NMT-LM & 75.0 / 59.4 & 56.0 & 62.4 & 26.9 & 53.1  \\ \bottomrule      
\end{tabular}}
\caption{Results are averaged across all the languages in each dataset. We report F1/EM for QA, exact match accuracy (EM) for structured prediction, ROUGE-L \cite{Lin2004ROUGEAP} for summarization and F1 for NER. Each score is the median over five runs. The final columns lists the average of all the scores. Refer to Appendix~\ref{sec:all_task_results} for scores on individual languages.} \label{table:main}
\end{table*}


\paragraph{Hyperparameters}
Pre-training is done with a batch size of 1M tokens and fine-tuning with 131,072 tokens, with a constant learning rate of 0.001. 
Starting from publicly available mT5-Large checkpoints, we further pre-train for 100K steps with a mix of monolingual and parallel objectives. The parallel data is mixed into monolingual data at a 10\% ratio, which amounts to roughly 4 passes over the OPUS-100 corpus. Examples from each language pair are sampled using the same language sampling distribution as \citet{xue2020mT5}, with alpha=0.3. For downstream tasks, we fine-tune for 10K steps for TyDiQA, MTOP, NER and 25K for WikiLingua, since it is a much larger dataset. Checkpoint selection is done based on the validation set.

\paragraph{Baselines}
Our first baseline is the publicly available mT5-Large model (1.3 billion parameters). For a fair comparison, we also experiment with an mT5 model further pre-trained for 100k steps with \textit{only} monolingual data from mC4 (see row 2: mT5+MLM in Table \ref{table:main}). This lets us assess whether improvements stem from using parallel data or just pre-training for longer. 

\section{Results}
We report results in table \ref{table:main}. Overall, adding parallel data through neural machine translation objectives improves scores for all 4 tasks, with the NMT objective performing the best. 
\par Simply pre-training mT5 for longer with just monolingual data (MLM) leads to improved scores for all tasks. The TLM objective is not be able to effectively leverage the parallel data and performs on par with MLM. On the other hand, our three NMT-based objectives show gains over MLM across all tasks. Among these, NMT and Denoised-NMT are the best and perform similarly, while Denoised-NMT+LM fares slightly worse. Averaged across all tasks, NMT and Denoised-NMT outperform MLM by 4 points.

\subsection{Model size}
\citet{xue2020mT5} find that cross-lingual performance of language models increases monotonically with model size. To study the impact of model capacity, we also experiment with larger model sizes. Even at the XL size (3.7B params, 3$\times$ larger than mT5-Large), we observe gains for all tasks with nmT5 (Table \ref{tab:size}). However, the magnitude of the gains is largely diminished, hinting that the need for parallel data reduces as model capacity increases. This finding is particularly promising for low-resource languages, where it is difficult to obtain high-quality parallel data.
\par At the same time, nmT5-Large substantially reduces the performance gap between mT5-Large and mT5-XL, covering 70\% of the headroom. Since bigger models are expensive to train and even more expensive to deploy, this opens up avenues for effectively using parallel data to improve performance of smaller language models. \citet{Turc2019WellReadSL} found that pre-training student models before model distillation is helpful, and using parallel data to improve student pre-training is another interesting avenue of future work.

\begin{table}[!htb]
\resizebox{0.5\textwidth}{!}{\begin{tabular}{l|ccccc} \toprule
Model           & TyDi QA     & MTOP & NER & WikiLingua & Avg.  \\ \midrule
mT5-Large  & 66.3 / 49.8 & 43.7 & 58.4 & 25.2 & 46.3      \\
nmT5-Large & 75.1 / 60.1 & 57.7 & 61.4  & 27.4  & 53.5      \\
$\Delta$ & 8.8 / 10.3 & 14.0 &  3.0 & 2.2 &  7.2 \\
\midrule
mT5-XL     & 77.8 / 61.8 & 63.4 & 65.5  & 27.9  & 56.7    \\
nmT5-XL    & 78.4 / 63.3 & 64.9 & 66.2  & 28.4 & 57.6      \\
$\Delta$ & 0.6 / 1.5 & 1.5 & 0.7 & 0.5 & 0.9 \\
\bottomrule
\end{tabular}}
\caption{Impact of model size on nmT5's performance.}\label{tab:size}
\end{table}

 
\subsection{Limited labeled data}
The TyDi QA dataset has only 3.7K English training examples. To study the impact of the size of fine-tuning data, we run experiments in two additional settings: a \textit{few-shot} regime and a \textit{high data} regime. Few-shot uses just 100 randomly sampled training examples, while for the latter we use the much larger SQuAD corpus \cite{Rajpurkar2016SQuAD10}, which consists of 80k examples. 
\par When fine-tuned with SQuAD, nmT5 performs slightly better than mT5 for both Large and XL model sizes. However, in the few-shot setting, nmT5-Large improves over mT5-Large by 15 points. Even at the XL size, nmT5 is over 10 points higher than mT5. nmT5-Large even outperforms the much larger mT5-XL. Our experiments suggest that pre-training with parallel data is particularly useful in the limited labelled data setting. 

\begin{table}[]
\resizebox{0.48\textwidth}{!}{\begin{tabular}{l|lcc} \toprule
Model      & Few-Shot (100) & Low (3.7K)  & High (80K)  \\ \midrule
mT5-Large  & 33.1 / 23.6    & 66.3 / 49.8 & 78.1 / 64.8 \\
nmT5-Large & 48.8 / 37.1    & 75.1 / 60.1 & 78.2 / 65.5 \\
$\Delta$ & 15.7 / 13.5 & 8.8 / 10.3 & 0.1 / 0.7\\
\midrule
mT5-XL  & 45.0 / 31.7 & 77.8 / 61.8 & 78.7 / 65.8 \\
nmT5-XL & 57.2 / 44.4 & 78.4 / 63.3 & 79.7 / 67.0 \\ 
$\Delta$ & 12.2 / 12.7 & 0.6 / 1.5 & 1.0 / 1.2
\\
\bottomrule
\end{tabular}} \label{tab:tydiqa}
\caption{Performance on the TyDi QA eval set when fine-tuned in the \textit{few-shot} (100 examples from TyDi QA English), \textit{low}  (full TyDi QA English with 3.7K examples) and \textit{high} data regime (SQuAD English with 80K examples).}
\end{table}

\subsection{Mixing ratio}
So far, we have mixed parallel data into monolingual data at a 10\% ratio. To assess how the mixing ratio impacts performance, we compare results with a 50\% mix. With the 50\% mix, average performance is slightly lower, validating our initial choice.
\begin{table}[h]
\resizebox{0.48\textwidth}{!}{\begin{tabular}{l|ccccc}  \toprule
Mix  & TyDi QA     & MTOP & NER & WikiLingua & Avg. \\ \midrule
10\% & 75.1 / 60.1 & 57.7 & 61.4 & 27.4 & 53.5 \\
50\% & 76.5 / 60.1 & 53.9 & 62.0 & 26.5 & 52.7     \\  \bottomrule
\end{tabular}}
\caption{Impact of mixing ratio on nmT5.} \label{tab:mixingratio}
\end{table}

\vspace{-4mm}

\subsection{Performance on unseen languages} \label{sec:unseen}
We also test downstream performance on languages previously unseen by the models. We randomly pick 30 languages from the WikiAnn NER dataset that are not covered in either mC4 \footnote{Subject to precision of language ID models used for mC4.} or OPUS, and hence none of our models have seen them during pre-training. Table \ref{tab:unseen} shows nmT5 outperforms mT5 on this subset of languages as well, indicating that the representations of the nmT5 model are better suited for cross-lingual transfer.

\begin{table}[h]
\centering
\resizebox{0.4\textwidth}{!}{
\begin{tabular}{l|ccc|c}
\toprule
 Model    & ckb  & hsb  & xmf  & ``Avg.'' \\
\midrule
mT5-Large     & 66.5 & 64.8 & 58.4 & 54.9        \\
nmT5-Large & 72.2 & 69.8 & 62.2 & 57.4       \\
$\Delta$ & 5.7 & 5.0 & 3.8 & 2.5 \\
\bottomrule
\end{tabular}
}
\caption{Performance on three randomly picked unseen languages. ``Avg.'' is calculated by averaging performance across 30 unseen languages.} \label{tab:unseen}
\end{table} 

\section{Related Work}

Pre-trained multilingual models such as mBERT and XLM-R have shown to be effective at cross-lingual transfer learning \cite{devlin-etal-2019-bert,conneau2020unsupervised}. 
Subsequently, many attempts have leveraged parallel data to improve cross-lingual capability of these models. 
\citet{conneau2019advances} proposed translation language modeling (TLM), to encourage the model to align representations across languages. 
Alternating language modeling \cite{yang2020alternating} and back-translation masked language modeling \cite{ouyang2020ernie} used code-switched sentences and back-translation respectively to utilize parallel data. Other works using parallel data in this line of work include FILTER \cite{fang2020filter}, AMBER \cite{hu2020explicit} and, MMTE \cite{siddhant2020evaluating}. A key factor that differentiates this paper from these works is that our pre-trained models use a text-to-text architecture, having both an encoder and a decoder, while the aforementioned models only have the encoder. Other pretrained multilingual encoder-decoder models such as mT5 \cite{xue2020mT5}, mBART \cite{liu2020multilingual} and MASS \cite{song2019mass} do not make use of parallel data during pre-training.

\section{Conclusion}
In this work we attempted to improve mT5 pre-training by incorporating parallel data. We experimented with various text-to-text objectives and found that  multi-tasking with the standard neural machine translation objective during pre-training leads to improved cross-lingual transfer. The improvements from parallel data are most pronounced in the limited labeled data scenario. Our experiments also indicate that smaller models, with the help of parallel data, can approach the performance of larger ones, while also suggesting that the need for parallel data is lesser as the model capacity increases.

\bibliographystyle{acl_natbib}
\bibliography{anthology,acl2021}

\clearpage

\onecolumn

\appendix

\input{appendix}

\end{document}

%% file: appendix.tex
\section{Per-Language Results on All Tasks}
\label{sec:all_task_results}
\begin{table*}[h!]
\centering
\begin{tabular}{l|ccccc}
\toprule
                     & \multicolumn{1}{c}{en} & \multicolumn{1}{c}{ar} & \multicolumn{1}{c}{bn} & \multicolumn{1}{c}{fi} & \multicolumn{1}{c}{id}  \\
                     \midrule
mt5                  & 75.0 / 63.0            & 68.9 / 51.4            & 54.5 / 37.2            & 70.4 / 54.6            & 74.3 / 57.0             \\
+MLM                 & 78.5 / 68.2            & 76.1 / 59.9            & 59.0 / 40.7            & 73.5 / 61.0            & 76.7 / 60.0             \\
+MLM+TLM             & 77.3 / 67.0            & 75.7 / 57.2            & 61.7 / 39.8            & 73.3 / 59.0            & 77.0 / 60.0             \\
+MLM+NMT             & 78.4 / 69.3            & 78.9 / 63.1            & 74.0 / 54.9            & 77.0 / 64.8            & 79.9 / 64.8             \\
+MLM+denoised NMT    & 78.7 / 68.6            & 79.8 / 64.7            & 72.6 / 53.1            & 77.2 / 64.2            & 79.8 / 67.6             \\
+MLM+denoised NMT-LM & 78.2 / 68.2            & 78.8 / 62.3            & 69.1 / 49.6            & 78.2 / 65.7            & 79.6 / 64.8             \\
\midrule
                     & \multicolumn{1}{c}{ko} & \multicolumn{1}{c}{ru} & \multicolumn{1}{c}{sw} & \multicolumn{1}{c}{te} & \multicolumn{1}{c}{avg} \\
                     \midrule
mt5                  & 57.4 / 47.5            & 61.5 / 37.1            & 69.7 / 52.5            & 65.5 / 48.0            & 66.3 / 49.8             \\
+MLM                 & 64.4 / 55.4            & 68.6 / 48.9            & 74.2 / 57.7            & 71.1 / 48.6            & 71.3 / 55.6             \\
+MLM+TLM             & 66.5 / 55.8            & 67.8 / 48.0            & 73.9 / 57.1            & 66.5 / 47.5            & 71.1 / 54.6             \\
+MLM+NMT             & 64.9 / 56.2            & 72.1 / 51.8            & 77.2 / 63.1            & 73.3 / 53.1            & 75.1 / 60.1             \\
+MLM+denoised NMT    & 67.9 / 58.7            & 71.9 / 51.5            & 75.7 / 59.7            & 74.3 / 53.5            & 75.3 / 60.2             \\
+MLM+denoised NMT-LM & 67.8 / 59.4            & 72.7 / 51.1            & 76.0 / 59.9            & 74.4 / 54.0            & 75.0 / 59.4  \\
\bottomrule
\end{tabular}
\caption{TyDi QA GoldP results (F1/EM) for each language.} \label{tab:tydiqaresults}
\end{table*}

\begin{table*}[h!]
\centering
\begin{tabular}{l|ccccccc}
\toprule
                     & \multicolumn{1}{l}{en} & \multicolumn{1}{l}{de} & \multicolumn{1}{l}{es} & \multicolumn{1}{l}{fr} & \multicolumn{1}{l}{hi} & \multicolumn{1}{l}{th} & \multicolumn{1}{l}{avg} \\
                     \midrule
mt5                  & 83.5                   & 41.2                   & 45.4                   & 43.3                   & 21.3                   & 27.5                   & 43.7                    \\
+MLM                 & 83.3                   & 44.5                   & 46.3                   & 51.8                   & 31.9                   & 34.0                   & 48.6                    \\
+MLM+TLM             & 85.0                   & 42.4                   & 47.5                   & 49.6                   & 31.8                   & 35.2                   & 48.6                    \\
+MLM+NMT             & 86.1                   & 55.1                   & 59.0                   & 61.7                   & 42.2                   & 42.1                   & 57.7                    \\
+MLM+denoised NMT    & 85.8                   & 51.6                   & 55.2                   & 59.5                   & 42.7                   & 43.9                   & 56.5                    \\
+MLM+denoised NMT-LM & 85.9                   & 51.9                   & 55.0                   & 57.0                   & 44.1                   & 41.9                   & 56.0      \\
\bottomrule
\end{tabular}
\caption{MTOP results (EM) for each language.} \label{tab:mtopresults}
\end{table*}

\clearpage

\begin{table*}
\centering
\resizebox{\textwidth}{!}{\begin{tabular}{l|cccccccccccccc}
\toprule
                     & \multicolumn{1}{l}{en} & \multicolumn{1}{l}{af} & \multicolumn{1}{l}{ar} & \multicolumn{1}{l}{bg} & \multicolumn{1}{l}{bn} & \multicolumn{1}{l}{de} & \multicolumn{1}{l}{el} & \multicolumn{1}{l}{es} & \multicolumn{1}{l}{et} & \multicolumn{1}{l}{eu} & \multicolumn{1}{l}{fa} & \multicolumn{1}{l}{fi} & \multicolumn{1}{l}{fr}  & \multicolumn{1}{l}{he}        \\
                     \midrule
mt5                  & 80.5                   & 64.5                   & 47.7                   & 57.2                   & 66.5                   & 67.0                   & 63.9                   & 62.0                   & 59.0                   & 45.5                   & 41.4                   & 56.9                   & 76.7                    & 45.1                          \\
+MLM                 & 81.4                   & 65.1                   & 50.2                   & 55.2                   & 69.3                   & 68.6                   & 66.9                   & 70.5                   & 62.8                   & 46.6                   & 44.9                   & 58.9                   & 76.6                    & 46.4                          \\
+MLM+TLM             & 82.4                   & 65.6                   & 48.8                   & 67.2                   & 72.2                   & 70.1                   & 70.8                   & 72.6                   & 61.2                   & 47.5                   & 47.1                   & 61.4                   & 78.7                    & 48.0                          \\
+MLM+NMT             & 82.2                   & 64.2                   & 56.7                   & 61.0                   & 69.1                   & 70.5                   & 64.6                   & 66.3                   & 66.2                   & 49.3                   & 48.9                   & 60.6                   & 78.4                    & 46.2                          \\
+MLM+denoised NMT    & 82.5                   & 65.7                   & 50.3                   & 63.6                   & 69.6                   & 70.7                   & 68.6                   & 73.7                   & 64.9                   & 48.6                   & 44.3                   & 63.3                   & 77.7                    & 45.5                          \\
+MLM+denoised NMT-LM & 82.9                   & 66.1                   & 49.5                   & 67.7                   & 74.5                   & 71.1                   & 71.3                   & 74.2                   & 67.1                   & 49.9                   & 44.8                   & 63.2                   & 80.2                    & 49.6                          \\
\midrule
                     & \multicolumn{1}{l}{hi} & \multicolumn{1}{l}{hu} & \multicolumn{1}{l}{id} & \multicolumn{1}{l}{it} & \multicolumn{1}{l}{ja} & \multicolumn{1}{l}{jv} & \multicolumn{1}{l}{ka} & \multicolumn{1}{l}{kk} & \multicolumn{1}{l}{ko} & \multicolumn{1}{l}{ml} & \multicolumn{1}{l}{mr} & \multicolumn{1}{l}{ms} & \multicolumn{1}{l}{my}  & \multicolumn{1}{l}{nl}        \\
                     \midrule
mt5                  & 66.8                   & 57.7                   & 44.9                   & 75.4                   & 36.0                   & 46.0                   & 53.0                   & 22.5                   & 29.5                   & 44.8                   & 38.6                   & 65.5                   & 27.0                    & 77.3                          \\
+MLM                 & 66.5                   & 61.4                   & 46.2                   & 76.4                   & 35.8                   & 49.0                   & 53.6                   & 23.7                   & 31.4                   & 46.0                   & 39.3                   & 67.4                   & 33.0                    & 78.5                          \\
+MLM+TLM             & 69.6                   & 61.9                   & 47.2                   & 76.7                   & 37.3                   & 51.0                   & 59.4                   & 29.3                   & 30.7                   & 48.2                   & 42.1                   & 70.2                   & 29.0                    & 80.4                          \\
+MLM+NMT             & 69.8                   & 61.7                   & 46.1                   & 77.3                   & 34.5                   & 53.0                   & 55.2                   & 27.0                   & 31.4                   & 43.0                   & 46.7                   & 69.0                   & 27.0                    & 78.9                          \\
+MLM+denoised NMT    & 65.8                   & 63.0                   & 46.6                   & 77.6                   & 37.0                   & 54.0                   & 58.3                   & 26.4                   & 29.8                   & 44.8                   & 42.1                   & 64.3                   & 30.0                    & 80.2                          \\
+MLM+denoised NMT-LM & 67.7                   & 64.4                   & 48.1                   & 77.9                   & 39.2                   & 49.0                   & 59.4                   & 30.0                   & 31.4                   & 47.4                   & 36.4                   & 71.0                   & 34.0                    & 80.2                          \\
\midrule
                     & \multicolumn{1}{l}{pt} & \multicolumn{1}{l}{ru} & \multicolumn{1}{l}{sw} & \multicolumn{1}{l}{ta} & \multicolumn{1}{l}{te} & \multicolumn{1}{l}{th} & \multicolumn{1}{l}{tl} & \multicolumn{1}{l}{tr} & \multicolumn{1}{l}{ur} & \multicolumn{1}{l}{vi} & \multicolumn{1}{l}{yo} & \multicolumn{1}{l}{zh} & \multicolumn{1}{l}{avg} & \multicolumn{1}{l}{\textbf{}} \\
                     \midrule
mt5                  & 73.1                   & 48.4                   & 66.8                   & 39.9                   & 37.9                   & 8.5                    & 77.8                   & 57.6                   & 45.1                   & 76.4                   & 58.0                   & 41.8                   & 58.4                    & \multicolumn{1}{l}{\textbf{}} \\
+MLM                 & 75.5                   & 47.3                   & 64.5                   & 40.5                   & 38.0                   & 9.2                    & 76.9                   & 56.5                   & 51.7                   & 76.9                   & 59.0                   & 41.8                   & 59.9                    & \multicolumn{1}{l}{\textbf{}} \\
+MLM+TLM             & 76.3                   & 58.8                   & 66.3                   & 40.2                   & 41.2                   & 8.8                    & 76.9                   & 62.0                   & 43.0                   & 79.6                   & 56.0                   & 43.5                   & 61.4                    & \multicolumn{1}{l}{\textbf{}} \\
+MLM+NMT             & 75.5                   & 56.0                   & 65.8                   & 40.3                   & 41.6                   & 8.0                    & 78.7                   & 60.3                   & 57.0                   & 79.8                   & 63.0                   & 41.0                   & 61.4                    & \multicolumn{1}{l}{\textbf{}} \\
+MLM+denoised NMT    & 75.5                   & 58.9                   & 66.2                   & 40.4                   & 40.4                   & 7.9                    & 78.7                   & 60.5                   & 50.0                   & 80.3                   & 64.0                   & 41.4                   & 61.5                    & \multicolumn{1}{l}{\textbf{}} \\
+MLM+denoised NMT-LM & 78.6                   & 60.9                   & 65.6                   & 40.6                   & 40.9                   & 9.1                    & 77.0                   & 63.1                   & 53.5                   & 79.8                   & 60.0                   & 45.5                   & 62.4                    & \multicolumn{1}{l}{\textbf{}}
\\
\bottomrule
\end{tabular}}
\caption{WikiAnn NER results (F1) for each language.} \label{tab:nerresults}
\end{table*}

\begin{table*}[]
\centering
\begin{tabular}{l|ccccccccccc}
\toprule
                    & en   & ar   & cs   & de   & es   & fr   & hi   & id   & it   & ja        \\
                    \midrule
mt5                  & 29.2 & 23.2 & 22.4 & 25.0 & 25.3 & 24.6 & 25.2 & 25.3 & 24.1 & 26.2      \\
+MLM                 & 30.0 & 24.0 & 22.9 & 26.0 & 26.6 & 25.5 & 26.1 & 25.8 & 24.9 & 27.8      \\
+MLM+TLM             & 30.0 & 24.4 & 23.1 & 25.6 & 26.3 & 25.6 & 26.4 & 25.8 & 25.1 & 27.6      \\
+MLM+NMT             & 31.5 & 25.7 & 24.0 & 27.0 & 27.5 & 26.4 & 27.7 & 27.0 & 25.8 & 29.5      \\
+MLM+denoised NMT    & 31.3 & 25.7 & 24.7 & 27.3 & 27.5 & 26.8 & 27.8 & 27.2 & 25.8 & 29.2      \\
+MLM+denoised NMT-LM & 30.8 & 25.0 & 23.7 & 26.5 & 27.1 & 26.3 & 27.3 & 26.7 & 25.6 & 28.7      \\
\midrule
                     & ko   & nl   & pt   & ru   & th   & tr   & vi   & zh   & avg  & \textbf{} \\
                     \midrule
mt5                  & 23.8 & 25.7 & 24.6 & 23.9 & 25.3 & 30.9 & 22.9 & 25.8 & 25.2 & \textbf{} \\
+MLM                 & 25.2 & 26.5 & 25.3 & 24.6 & 27.1 & 31.1 & 23.2 & 27.1 & 26.1 & \textbf{} \\
+MLM+TLM             & 24.7 & 26.6 & 25.2 & 24.4 & 26.5 & 31.3 & 23.3 & 27.0 & 26.1 & \textbf{} \\
+MLM+NMT             & 26.7 & 27.7 & 26.3 & 25.9 & 28.6 & 34.1 & 23.9 & 28.1 & 27.4 & \textbf{} \\
+MLM+denoised NMT    & 26.6 & 28.0 & 25.9 & 25.8 & 28.3 & 33.4 & 24.3 & 28.4 & 27.4 & \textbf{} \\
+MLM+denoised NMT-LM & 25.9 & 27.4 & 25.6 & 24.9 & 27.3 & 33.1 & 23.8 & 27.8 & 26.9 & \textbf{} \\
\bottomrule
\end{tabular}
\caption{Wikilingua results (Rouge-L) for each language.} \label{tab:wikilinguaresults}
\end{table*}

\clearpage

\begin{table*}[]
\centering
\begin{tabular}{l|ccccccccccc}
\toprule
           & ace  & arz  & ast  & ba   & ce   & ckb  & csb  & eml  & fur  & gan       & gn   \\
           \midrule
mt5-Large  & 44.8 & 50.8 & 83.3 & 38.1 & 21.7 & 66.5 & 56.7 & 39.8 & 64.2 & 42.1      & 48.2 \\
nmt5-Large & 46.7 & 53.6 & 84.8 & 43.7 & 28.3 & 72.2 & 58.1 & 41.9 & 65.6 & 41.2      & 51.0 \\
\midrule
           & hsb  & ia   & jbo  & lij  & lmo  & min  & nap  & nov  & pdc  & pms       & pnb  \\
           \midrule
mt5-Large  & 64.8 & 63.2 & 42.1 & 46.3 & 69.8 & 39.1 & 62.2 & 62.1 & 48.1 & 81.5      & 61.1 \\
nmt5-Large & 69.8 & 62.4 & 43.6 & 43.0 & 72.0 & 45.5 & 61.7 & 66.7 & 51.2 & 83.5      & 55.4 \\
\midrule
           & rm   & sa   & tl   & qu   & vec  & vep  & vls  & xmf  & avg  &           &      \\
           \midrule
mt5-Large  & 64.1 & 17.4 & 78.6 & 27.5 & 66.9 & 63.6 & 74.4 & 58.4 & 54.9 & \textbf{} &      \\
nmt5-Large & 67.6 & 23.0 & 79.4 & 35.6 & 66.7 & 68.0 & 77.5 & 62.2 & 57.4 & \textbf{} &     
 \\
\bottomrule
\end{tabular}
\caption{WikiAnn NER results on unseen languages. Refer to section \ref{sec:unseen}} \label{tab:wikilinguaresults}
\end{table*}